\documentclass[10pt,letterpaper]{article}
\usepackage[top=0.85in,left=2.75in,footskip=0.75in]{geometry}

\usepackage{amsmath,amssymb}

\usepackage{changepage}

\usepackage{textcomp,marvosym}

\usepackage{cite}

\usepackage{nameref,hyperref}

\usepackage[nopatch=eqnum]{microtype}
\DisableLigatures[f]{encoding = *, family = * }

\usepackage[table]{xcolor}

\usepackage{array}

\usepackage{graphicx}
\usepackage{times}
\usepackage{latexsym}
\usepackage{amsmath} 
\usepackage{multirow}
\usepackage{rotating}
\usepackage{url}
\usepackage{subcaption}
\usepackage[numbers,sort&compress]{natbib}
\usepackage{algorithm}
\usepackage{algpseudocode}
\usepackage{paralist}
\usepackage{acronym}
\usepackage{mathtools}
\usepackage{helvet}
\usepackage{courier}
\usepackage{enumitem}
\usepackage{algpseudocode}
\usepackage{subcaption}
\usepackage{makecell}
\usepackage[skip=0.333\baselineskip]{caption}

\newcolumntype{+}{!{\vrule width 2pt}}

\newlength\savedwidth

\raggedright
\usepackage[aboveskip=1pt,labelfont=bf,labelsep=period,justification=raggedright,singlelinecheck=off]{caption}

\makeatletter
\renewcommand{\@biblabel}[1]{\quad#1.}
\makeatother

\usepackage{lastpage,fancyhdr,graphicx}
\usepackage{epstopdf}
\fancyheadoffset[L]{2.25in}
\fancyfootoffset[L]{2.25in}
\begin{document}
\vspace*{0.2in}

\begin{flushleft}
{\Large
\textbf\newline{Inferring gender from name: a large scale performance evaluation study} 
}
\newline
\\
Kriste Krstovski\textsuperscript{1,2*},
Yao Lu \textsuperscript{2,3},
Ye Xu\textsuperscript{3}

\bigskip
\textbf{1} Columbia Business School, Columbia University, New York, NY, USA
\\
\textbf{2} Data Science Institute, Columbia University, New York, NY, USA 
\\
\textbf{3} Department of Sociology, Columbia University, New York, NY, USA 
\\
\bigskip

* kriste.krstovski@columbia.edu

\end{flushleft}
\section*{Abstract}
A person's gender is a crucial piece of information when performing research across a wide range of scientific disciplines, such as medicine, sociology, political science, and economics, to name a few. However, in increasing instances, especially given the proliferation of big data, gender information is not readily available. In such cases researchers need to infer gender from readily available information, primarily from persons' names. While inferring gender from name may raise some ethical questions, the lack of viable alternatives means that researchers have to resort to such approaches when the goal justifies the means - in the majority of such studies the goal is to examine patterns and determinants of gender disparities. The necessity of name-to-gender inference has generated an ever-growing domain of algorithmic approaches and software products. These approaches have been used throughout the world in academia, industry, governmental and non-governmental organizations. Nevertheless, the existing approaches have yet to be systematically evaluated and compared, making it challenging to determine the optimal approach for future research. In this work, we conducted a large scale performance evaluation of existing approaches for name-to-gender inference. Analysis are performed using a variety of large annotated datasets of names. We further propose two new hybrid approaches that achieve better performance than any single existing approach.


\section*{Introduction}
\label{sec:introduction}
The proliferation of big data has spurred researchers to harness its potential, utilizing its immense volume of information to understand and address societal issues, such as those related to gender inequality. For example, health professionals seek to understand how certain diseases and treatments deferentially affect women compared to men \cite{kautzkywiller_ea_2016,ward_ea_2004}. Sociologists strive to understand the contributing factors for the persistent gender gap in the economic domain \cite{england_ea_2020}, and education \cite{Kao_2015}. Political scientists are motivated to study the gender gap in various forms of political participation \cite{robinson_gottlieb_2021} and voter turnout \cite{kostelka_ea_2019}. Economists study the gender gap in the labour market \cite{altonji_ea_1999} and the ethnic capital \cite{borjas_2021}. These examples are just a mere glimpse into the immense range of possibilities that emerge when big data coupled with gender information could profoundly shape our daily lives and narrow the persisting gender gap that permeates our societies. 

Studies of this nature require feature-rich information on the individual level such as detailed information about the patient, resident, employee, voter or community member. However, it is often the case that individuals' personally identifiable information, such as gender, is either missing or not readily available in big data (e.g., gender is not directly identified in resumes). Sometimes demographic information is retracted due to privacy concerns. The lack of demographic information significantly hampers the research potential of these data.

To account for the absence of gender information, various automated approaches for inferring person's gender from their names (first and/or last name) have been proposed in the past. These approaches have been widely adopted throughout the world by various leading academic institutions, non-government, and government organizations such as the United Nations and the European Commission \cite{namsor}. The available gender inference approaches vary from heuristic to probabilistic, and to neural network based. The existing approaches also vary in their input data, i.e. whether they only use a person's name or names supplemented with additional profile information \cite{zach, liu}. While many of them are free of charge, some are paid services with subscription plans. The wide selection of such approaches raises a natural question: how do these approaches perform and which provides the highest accuracy? The answer to this question has important implications for research.

One way to answer this question is by performing evaluations of the existing inference approaches. While previous studies have occasionally conducted evaluations, their evaluation is limited given the coverage of existing approaches and the reliance on small test sets. For example, in a previous study, several name-to-gender inference approaches have been evaluated on a set of only around 7,000 names \cite{santamaria_mihajlevic_2018}. 

To alleviate these limitations, we performed a large scale performance evaluation study of existing name-to-gender inference approaches. In our analysis we indexed most of the existing and readily available datasets of persons' names labeled with gender to construct large test sets of persons' names. We used these test sets and their combined version to evaluate and compare the performance of most of the existing name-to-gender inference approaches. Our test sets come from various sources such as voter registration, the Social Security Administration (SSA), and political contributors. Having a wide variety of test sets of persons' names allows us to perform a comprehensive performance evaluation study across a wide range of population groups. We evaluated a wide spectrum of existing inference approaches (free or subscription based). 

In addition to our performance evaluation, we introduce a new gender inference approach. We further present two new hybrid methods combining existing approaches with our new approach in order to obtain higher accuracy. We showcase that the proposed hybrid methods yield the highest accuracy across all test sets.

Our large scale performance evaluation of existing gender inference approaches involved curating available datasets of persons' names labeled with gender. Given that all such available datasets contain either male or female genders, we carry out the task as binary classification. 

This paper is organized as follows. We start by describing the datasets that we used in our evaluation study. In the subsequent sections, we describe the existing gender inference approaches and our new proposed approaches that combine different inference approaches based on voting. Finally we present the results from the performance evaluation study and error analysis which include measuring gender bias across all inference approaches.

\section*{Annotated Datasets}
\label{sec:gender_annotated_datasets}
We used six datasets of persons' names annotated with gender. They are: ACL Anthology gendered name dataset, CMU name dataset, DIME dataset, Facebook name data, Florida voter data, and SSA data. These datasets vary in terms of a person's name representation. They either contain person's first or full name (first and last name). Shown in Table~\ref{tab:name_gender_summary} are summary statistics over the six datasets. We provide the name type (first vs. full), total number of names ("Names"), number of unique names ("Unique Names"), number of unique first ("Unique First Names") and last names ("Unique Last Names"), and number of ambiguous names ("Ambiguous Names") which are names that appear as both female and male.

\begin{table}[!ht]
\begin{adjustwidth}{-2.25in}{0in} 
\centering
\caption{
{\bf Statistics over datasets of persons' names labeled with gender.}}
\begin{tabular}{|l|r|r|r|r|r|r|} \hline
\bf Dataset  &      \bf Name Type & \bf Names & \bf Unique Names & \bf Unique First Names & \bf Unique Last Names &  \bf Ambiguous Names \\ \hline
ACL      & Full       &       12,692      &       11,919 &              5,244 &        0 &               0 \\ \hline
CMU      & First      &        7,211      &        7,211 &              7,211 &        0 &               0 \\ \hline
DIME     & First      &   13,498,575      &      125,455 &            125,455 &        0 &          15,107 \\ \hline
Facebook & First      &      623,100      &       23,403 &             23,403 &        0 &           5,508 \\ \hline
Florida  & Full       &   13,357,053      &   12,308,570 &            433,104 &  884,756 &           29,351 \\ \hline
SSA      & First      &  421,456,434      &       63,892 &             63,892 &        0 &           7,050 \\ \hline
\end{tabular}
\label{tab:name_gender_summary}
\end{adjustwidth}
\end{table}

Prior to their use, we performed name cleaning across all datasets, which included removing names with only one letter in the name fields, those containing various personal titles, names without vowels, and names with unknown genders. In the following subsections we summarize each individual dataset.

\subsection*{ACL Anthology gendered name dataset}
\label{sec:gender_acl}
ACL Anthology is an archive of research papers published by the Association for Computational Linguistics (ACL). The associated ACL gendered name dataset was compiled as part of a paper examining gender differences in computational linguistic research topics \cite{vogel_ea_2012}. The dataset includes 12,692 ACL Anthology authors' full names and their genders. Names were annotated with gender labels using a variety of sources which include: 1990 U.S. Census data, manual annotation, and a website called Baby Name Guesser \cite{baby_name_guesser}, which provides information on the gender commonality of a baby name. There are 11,919 unique names in this collection, out of which 5,244 are unique first names and none of them are ambiguous. Given that in this dataset gender was inferred through an annotation process that involved statistics from existing datasets of names labeled with gender, we used this dataset only as a test set.

\subsection*{CMU name dataset}
\label{sec:gender_cmu}
The Carnegie Mellon University (CMU) name dataset \cite{cmu_names_data} consists of persons' first names created by Mark Kantrowitz. In this dataset, each name is not associated with the number of times it appears as male or female. Therefore, we used this dataset only as a test set in our performance evaluation. When creating the test set we removed ambiguous first names that were in both the female and male list. The resulting dataset contains 7,211 first names where all of them are unique and none are ambiguous.

\subsection*{Database on Ideology, Money in Politics, and Elections (DIME)}
\label{sec:gender_dime}
The database on Ideology, Money in Politics, and Elections (DIME) \cite{adam_2016} contains information on individual and institutional political contributors to local, state, and federal elections from 1979 to 2014. It was constructed for studying political finance, but the fact that it contains gender information of individual contributors makes it a useful resource for developing and evaluating name-to-gender inference approaches. The cleaned version of the DIME dataset has 13,498,575 first names, out of which 125,455 are unique first names and 15,107 are ambiguous.

\subsection*{Facebook Name Data}
\label{sec:gender_facebook}
The Facebook name data (Facebook) contains first names obtained from crawled data of Facebook user profiles from New York who have specified their gender \cite{tang_ea_2011}. What is unique about this dataset is that authors only included names with vowels (i.e. names without vowels were removed in the crawling process). The dataset was created with the end goal of analyzing gender differences in behaviors on the social media platform \cite{tang_ea_2011}. Using this labeled dataset, we created a gender frequency list for the crawled first names (i.e., for each first name, we included the frequency that name was associated with female vs. male users). The dataset consists of 623,100 first names, out of which 23,403 are unique first names and 5,508 are ambiguous.

\subsection*{Florida Data}
\label{sec:gender_florida}
The Florida dataset is created from the state of Florida official voting registration information. Per the Florida Division of Elections, ``Voter registration and voting history information is a public record under Florida law.`` \cite{florida_quote}. The official voting registration contains information on all voters registered or preregistered as of the end of the previous month, and includes relevant fields such as first name, last name, gender and race \cite{sood_laohaprapanon_2018}. We obtained the February 2017 Florida voting registration and history data from Harvard Dataverse \cite{crosas_2013}. The dataset contains 13,357,053 full names, out of which 12,308,570 are unique with 433,104 unique first names and 29,351 are ambiguous. 

\subsection*{SSA Data}
\label{sec:data_ssa}
The Social Security Administration (SSA) data is officially named "national data on the relative frequency of given names in the population of U.S. births where the individual has a social security number" \cite{ssa_2022}. On annual bases SSA creates a summary of the first names given to newborns. For each name we have the following information: first name, gender, and the number of occurrences of the first name with the given gender in that year. The dataset consists of annual summaries for the time period between 1870 and 2018. For privacy reasons SSA restricts the list of first names to those with at least five occurrences. We focused on the SSA data from 1937-1999. The resulting SSA data contains 421,456,434 first names out of which 63,892 are unique first names and 7,050 are ambiguous.

\subsection*{Gender Ambiguity}
\label{sec:data_gender_ambiguity}
As stated earlier, across the datasets, names may appear as both female and male. We refer to these names as gender ambiguous names. These names appear in datasets with full names (Florida) and datasets with first names only (DIME, Facebook, and SSA). We present in the last column of Table~\ref{tab:name_gender_summary} the total number of unique names that are gender ambiguous. For these names we counted the number of times they appear as female $\textnormal{count}(female)$ and as male $\textnormal{count}(male)$, respectively. We used these values to compute the maximum likelihood estimate (MLE) \cite{bishop_2006} of them being assigned as female: 
\begin{equation}
    MLE(female) = \frac{\textnormal{count}(female)}{\textnormal{count}(female)+\textnormal{count}(male)}
\label{eq:ml_gender}
\end{equation}

The MLE values allow us to estimate the probability of the name being female. The higher the value (close to 1), the more likely it is for the name to be female; vice versa, the closer the value is to 0, the more likely it is for the name to be male.

\begin{figure*}[h]
\captionsetup{justification=centering}
\begin{minipage}[t]{0.47\linewidth}
\begin{center}
\includegraphics[width=1\linewidth]{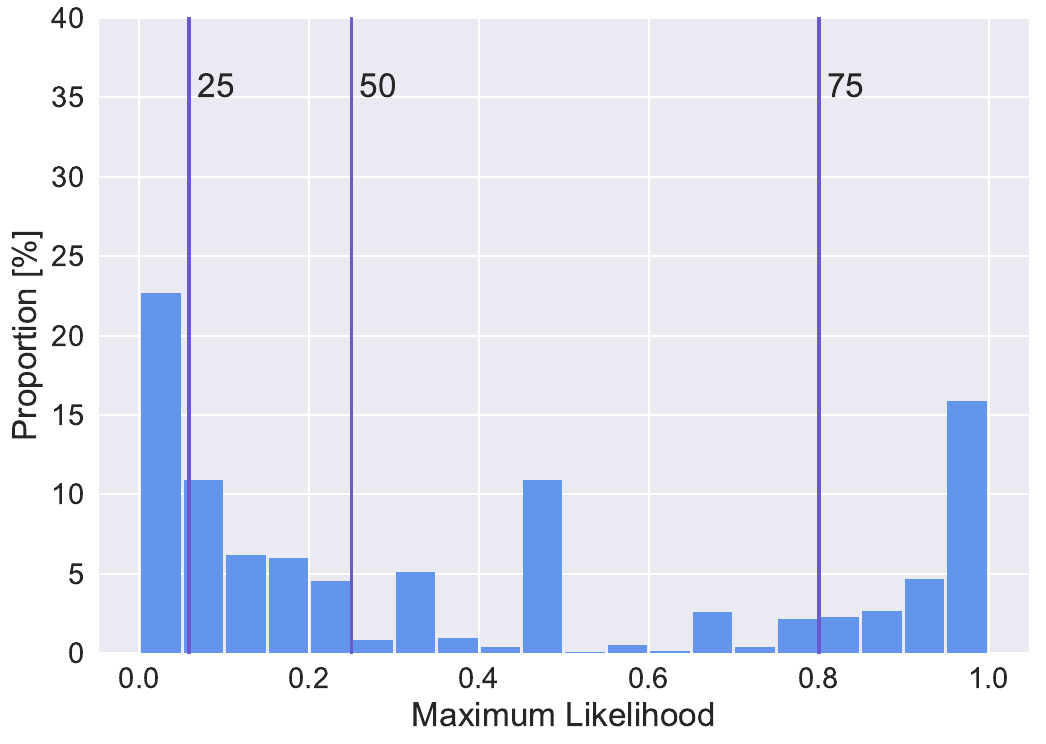} 
\subcaption{DIME}
\label{fig:dime}
\end{center} 
\end{minipage}
\hfill
\vspace{0.2 cm}
\begin{minipage}[t]{0.47\linewidth}
\begin{center}
\includegraphics[width=1\linewidth]{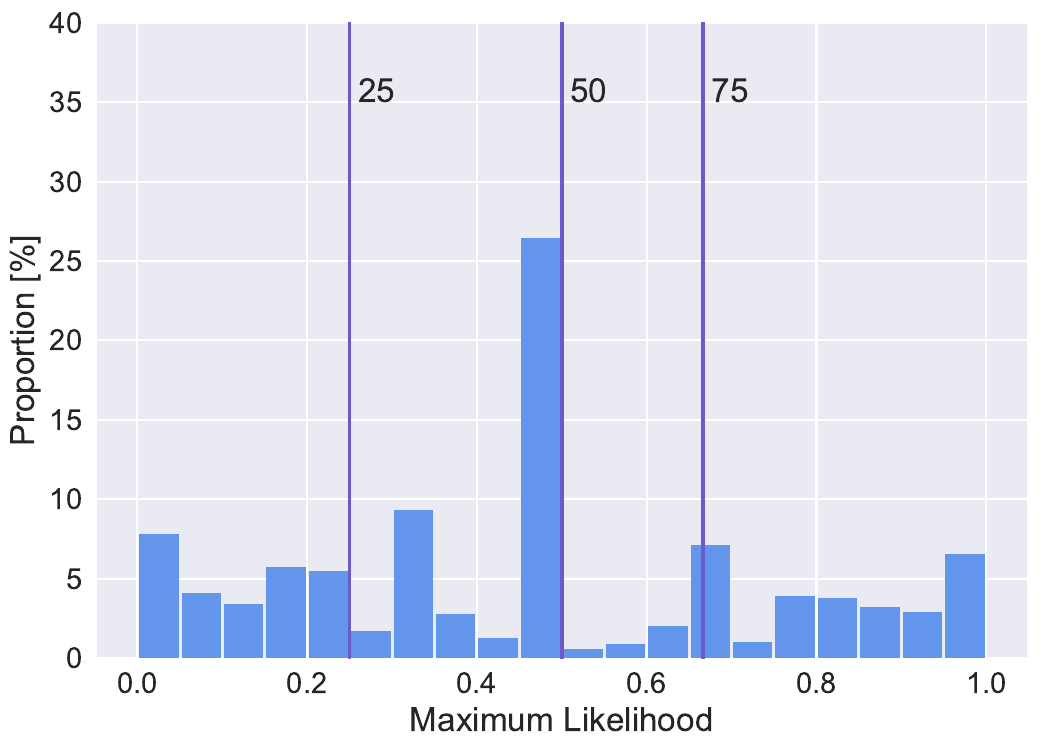} 
\subcaption{Facebook}
\label{fig:facebook}
\end{center}
\end{minipage}
\vfill
\vspace{0.2 cm}
\begin{minipage}[t]{0.47\linewidth}
\begin{center}
\includegraphics[width=1\linewidth]{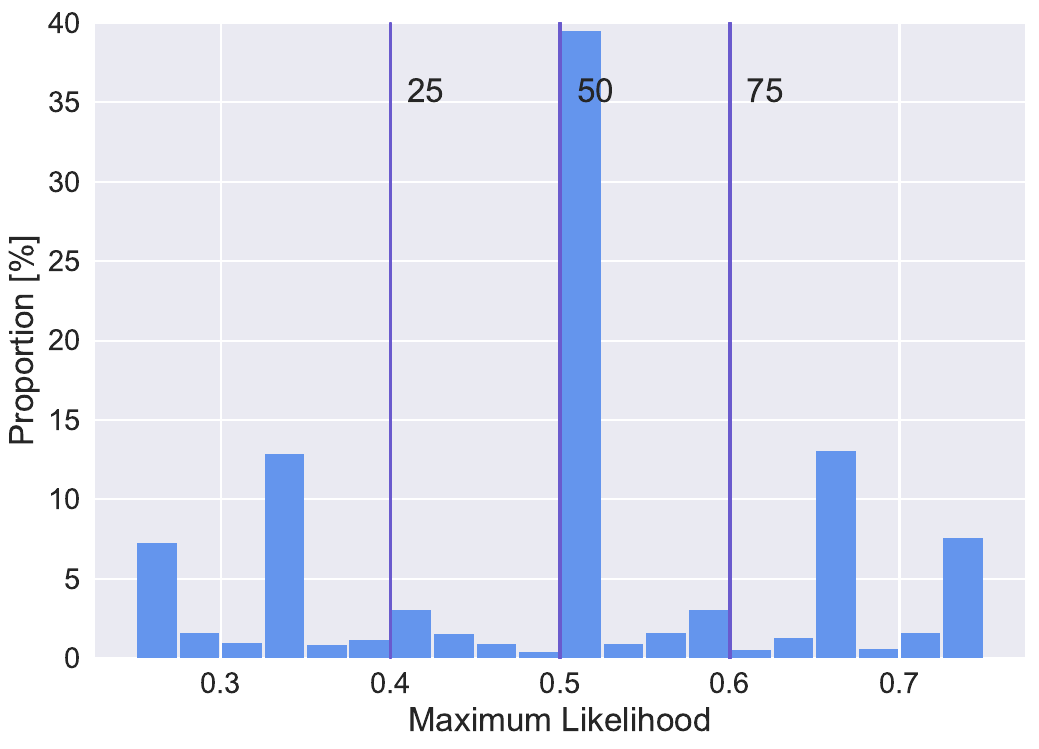} 
\subcaption{Florida}
\label{fig:florida}
\end{center}
\end{minipage}
\hfill
\begin{minipage}[t]{0.47\linewidth}
\begin{center}
\includegraphics[width=1\linewidth]{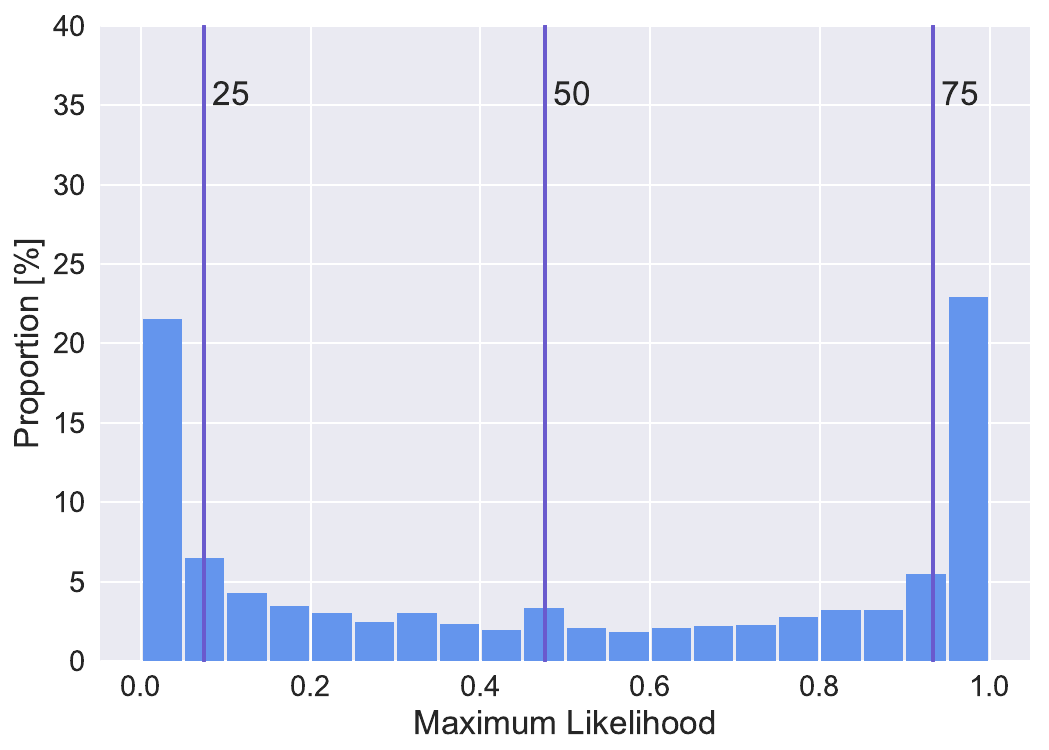} 
\subcaption{SSA}
\label{fig:ssa}
\end{center}
\end{minipage}
\caption{ML distribution of person's first name being female across ambiguous names within the four datasets. Violet lines indicate 25, 50, and 75 percentile of the ambiguous names}
\label{fig:mle_4datasets}
\end{figure*}

 In Figure~\ref{fig:mle_4datasets} we provide the distribution of the MLEs of the first name being female computed across the ambiguous names in the four datasets. The violet lines indicate the 25, 50, and 75 percentile of the ambiguous names.
 
 As we can see in the DIME and SSA datasets, the majority of the ambiguous names using MLE can be labeled as either male or female. For example, in the DIME dataset the 50\% line is at about 0.25, which means that about half of the ambiguous first names have ML values of less than 0.25; this indicates that the name is more likely to be male rather than female. A significant number of first names in this dataset have MLE value of exactly 0.5, which means that the first name has been used as male and female for exactly the same number of times. In most cases, this is because the name is used rarely. For example, a quick query shows that there are 1,489 names with only two occurrences, one as male and one as female. 
 
 In case of the Florida data we observe that the 50\% line is around 0.5. This means that after using MLE the majority of the ambiguous names remain ambiguous. We observe a similar distribution across the ambiguous names in the Florida data.

\subsection*{Name Uniqueness}
\label{sec:gender_name_uniqueness}
In Table~\ref{tab:unique_stats}, for each of the six datasets, we provide the number of dataset specific unique first names which are unique across all datasets ("Names Unique Across all Datasets"). In addition, we provide their percentage contribution in the dataset specific unique names ("\% of Dataset Specific Unique Names"). For example, out of the 125,455 unique first names found in the DIME collection, 53,082 of them are unique across all datasets (i.e. they can't be found in any of the other five datasets). These 53,082 unique names account for 42,31\% of the unique names in the DIME collection. Or in other words, 42,31\% of the unique names in the DIME collection are unique across all datasets. These statistics are important as they reveal how unique names are across the datasets in our study. We will return to these statistics when we discuss the performance analysis results, as they help distinguish how important the accuracy numbers are across the datasets. 

\begin{table}[t]
\centering
\begin{tabular}{|l|r|r|} \hline
\bf Dataset & \multicolumn{1}{c|}{\makecell{\bf Names Unique \\\bf  Across all Datasets}}  & \multicolumn{1}{c|}{\makecell{\bf  \% of Dataset Specific \\\bf  Unique Names }}\\
\hline
ACL      &   1,280 & 24.41\% \\ \hline
CMU      &     399 &  5.53\% \\ \hline
DIME     &  53,082 & 42.31\%  \\ \hline
Facebook &   1,550 &  6.62\% \\ \hline
Florida  & 341,749 & 78.91\% \\ \hline
SSA      &   4,626 &  7.24\% \\ \hline
\end{tabular}
\caption{Number and percentage of dataset specific unique first names that are unique across all datasets.}
\label{tab:unique_stats}
\end{table}

\subsection*{Combined Dataset}
\label{sec:gender_combined_dataset}
We also created a large dataset of names labeled with gender by combining four out of the six datasets. More specifically we used the DIME, Facebook, Florida, and the SSA data to create a larger dataset. For each name in the four datasets, we counted the number of times that the name appears as male and female. Data combination was done by adding the counts of male and female instances of each name. 

The purpose of creating this combined dataset was twofold. Firstly, we wanted to create the most comprehensive dataset of persons' names labeled with gender available. We believe that such a dataset would be very beneficial to the research community. Secondly, such a dataset would also facilitate the development of our own gender inference approach. 

When creating the combined dataset, as stated earlier, we left out the ACL and the majority of the CMU dataset. The ACL dataset was left out since gender was inferred from this dataset through an annotation process that involved statistics from existing datasets of names labeled with gender. We used this dataset only in our performance evaluation analysis. For the CMU dataset, given that all names are unique, adding all of them into the combined dataset would not affect the overall male and female count. However, out of the 7,211 first names in this dataset, we discovered that 399 of them are unique across all datasets. We therefore added only these unique names into our combined dataset in order to increase the name coverage.

\begin{figure}[!ht]
   \centering
   \includegraphics[width=0.47\linewidth]{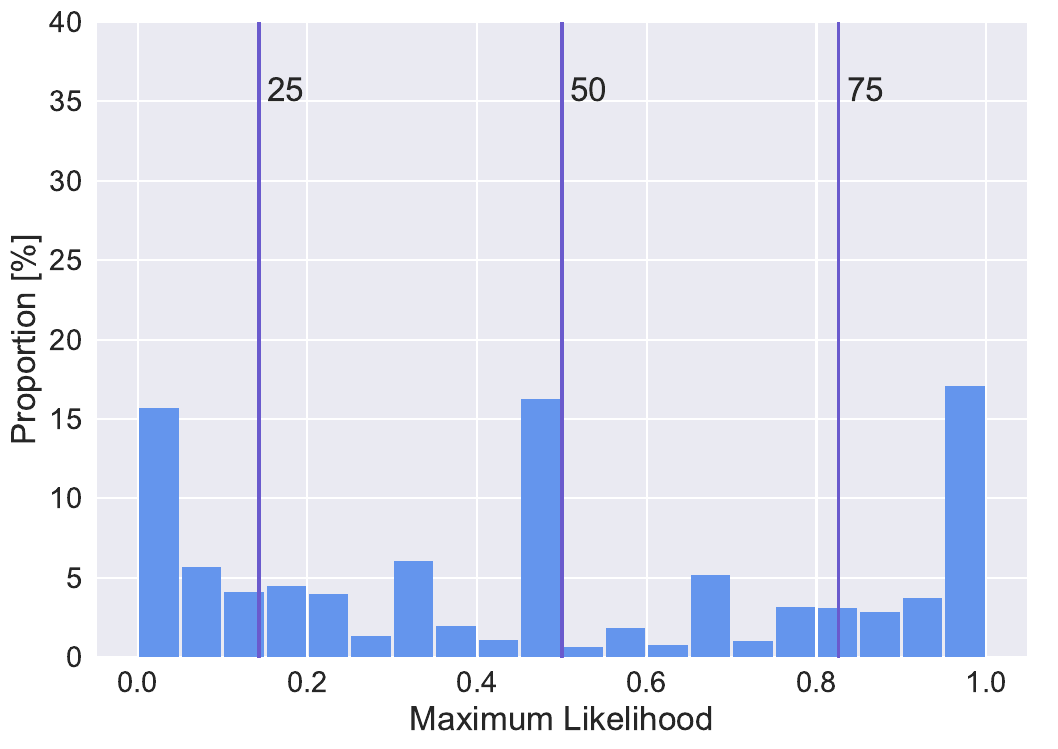}
   \caption{ML distribution (in \%) of the person's first name being female across the ambiguous names in the combined dataset. Violet lines indicate 25, 50, and 75 percentile of the ambiguous names.}
   \label{fig:mle_combined}
\end{figure}

The combined dataset consists of 494,927 unique first names, of which 44,582 are ambiguous. Fig~\ref{fig:mle_combined} provides the MLE distribution across the ambiguous first names in this combined dataset.

Unsurprisingly, the histogram combines the characteristics of the individual annotated datasets: there are three clusters of ML values, one around 0 (mostly male), one around 1 (mostly female), and one in the center (neutral).

To attain a better understanding of the coverage of the four datasets we computed the overlaps in unique first names between every pair of the datasets. Table~\ref{tab:dataset_pairs} shows the number of overlapping names between each pair of data. Diagonal entries are the number of unique first names for the given dataset. The value in parenthesis is the percentage of overlaps calculated as the number of overlapping names divided by the size of the dataset in that row. For example, 14,9\% in the first row and third column means that 14,9\% of the unique first names in the DIME data are also covered by the Facebook data.

\begin{table*}[t]
\centering
\begin{tabular}{|c|r|r|r|r|r|} \hline
\bf Dataset  & \bf DIME         & \bf Facebook    & \bf Florida     & \bf SSA         \\ \hline
    DIME     &  125,455         & 18,667 (14.9\%) & 70,290 (56.2\%) & 40,881 (32.6\%) \\ \hline
    Facebook &  18,667 (79.8\%) & 23,403          & 21,206 (90.6\%) & 16,247 (69.4\%) \\ \hline
    Florida  & 70,290 (16.2\%)  & 21,206 (4.9\%)  & 433,104         & 57,882 (13.4\%) \\ \hline
    SSA      & 40,881 (64.0\%)  & 16,247 (25.4\%) & 57,882 (90.6\%) & 63,892          \\ \hline
\end{tabular}
\caption{Number of overlapping unique first names between each pair of datasets and the percentage of dataset specific unique first names that they cover. For example, 13.4\% of the Florida unique names overlap with the SSA unique names.}
\label{tab:dataset_pairs}
\end{table*}

In each row, the percentages reflect how unique the dataset in that row is compared to the other datasets presented in the columns. The lower the percentages, the fewer first names that dataset has in common with other datasets. Comparing across each row shows that the Florida data contains the most unique first names among all four datasets, while Facebook is the least unique dataset.

\section*{Existing Name to Gender Inference Approaches}
\label{sec:gender_existing_approaches}
We analyzed the performance of six existing gender inference tools which, to the best of our knowledge, were available at the inception of our project. The existing gender inference tools are either accessible through an API (Baby Name Guesser, Gender API, and Namsor) or as a Python (chicksexer, gender-guesser) or R (gender-R) package. In addition, we also introduce our own gender inference approach.
 
In the following subsections, we give an overview of the six existing tools and describe our proposed approach. While we spent significant efforts in reviewing the documentation for each existing approach, we note that most of the approaches do not provide sufficient details about their algorithms due to their proprietary nature.

\subsection*{Baby Name Guesser}
\label{sec:gender_baby_name_guesser}
Baby Name Guesser \cite{baby_name_guesser} is an API that leveraged Google's Web Search SOAP API to analyze common patterns involving the first name. The API determines from popular usage whether the name was associated more frequently with male or female individuals. Using first name search data from the Google search engine, the developer claims to have amassed a database of over 100,000 first names. The API is accessed through the PHP name URL. No API key is needed to run this API. 

\subsection*{chicksexer}
\label{sec:gender_chicksexer}
Chicksexer is a Python package that infers gender from person's first or full name \cite{chicksexer}. It outputs the gender probability estimate. The inference is performed using a character-level Long Short-Term Memory (LSTM) neural network \cite{hochreiter_schmidhuber_1997}. The model is trained using three data sources: 1) SSA data, 2) dataset of first names retrieved from public records, and Dbpedia's person data \cite{dbpedia}, which is based on metadata from biographical Wikipedia articles.

\subsection*{Gender API}
\label{sec:gender_gender_api}
Gender API \cite{genderapi} is developed to infer gender from the person's first name, full name, or email address. If provided with a country code, it first searches the name in a specific country before doing a global lookup. If it cannot find the name in a global lookup, the API performs name cleaning and covers all spelling variants. For each predicted gender the API outputs a confidence score (on a scale from 0 to 100). Based on the information available on their website, the Gender API relies on a database of more than 42 million unique names originating from 191 different countries, six million of which have been validated \cite{genderapi}. Based on their documentation Gender API is continuously making efforts to improve their inference by analyzing more than ten million additional names on a monthly basis. The API usage is subscription based with a monthly quota.

\subsection*{gender-guesser}
\label{sec:gender_gender-guesser}
Gender-guesser is another Python package that performs gender prediction of a person's first name \cite{genderguesser}. Optionally users can provide the person's country as input. For each name the package outputs one of the following six categories: unknown, andy (androgynous), male, female, mostly male, or mostly female. Unknown is a label given to a name that is not in the database and andy is given when the name has the same probability of being male or female. The package performs the prediction using a dataset of over 40,000 first names labeled with gender and country of origin. The dataset claims to "cover the vast majority of first names in all European countries and in some overseas countries (e.g. China, India, Japan, U.S.A.) as well". 

\subsection*{Gender}
\label{sec:gender_gender-R}
Gender is an R package that infers gender using first name. Inference is based on historical data \cite{genderr}. In addition to the first name, the package allows the user to enter the specific birth year or range of years and the countries from which the names originate. The package outputs the predicted gender (male or female) along with the proportion of male and female names that were discovered in the package database for the specific year(s). When utilized, the user needs to specify the "method", namely the dataset to be used to predict the gender. The user can choose from five different methods: "ssa", "ipums", "napp", "kantrowitz", and "genderize". The "ssa" method relies on the SSA dataset. The "ipums" method relies on the U.S. Census data. The "napp" method uses data created by the North Atlantic Population Project, which consists of census microdata from Canada, Great Britain, Denmark, Iceland, Norway, and Sweden from 1801 to 1910. Lastly, the "kantrowitz" method uses a small dataset of male and female names. To avoid ambiguity we refer to this approach as Gender-R. Given that this approach consists of five different methods, which vary depending on the dataset used, in our study we evaluated all five of them and present the results of the best performing method which is the "SSA" based one.

\subsection*{Namsor}
\label{sec:gender_Namsor}
Namsor is an API that classifies a person's name by gender, country of origin, or ethnicity \cite{namsor}. It relies on sociolinguistics to extract semantics and determine the name origin. As for methodology, its website states that ``Our specialized data mining software recognizes the linguistic or cultural origin of personal names in any alphabet/language, with fine grain and high accuracy.`` \cite{namsor}. Namsor infers gender over person's first or full name. In addition, users can provide the most likely country of origin as an input. The API outputs the most likely gender along with its probability. The API additionally outputs a gender score on a scale from -1 (male) to 1 (female) and an unnormalized score. One of the advantages of this approach is that it covers names across a variety of national contexts. This is helpful in instances where a name's likely gender depends on language or national background. Namsor provides a comprehensive usage documentation. Use of Namsor requires paid subscription when going over a monthly quota of 5,000 names.

\section*{Maximum Likelihood Approach}
\label{sec:gender_ml_approach}
In addition to the existing gender inference approaches we developed a name-to-gender inference approach. This approach uses a straightforward method to assign gender to a person's name. It is based on the maximum likelihood estimates (MLE) of the probability of a person's name being female or male. MLEs are computed using the annotated datasets described in the previous section. We refer to this approach as the MLE. In the MLE approach for a given name, we first check to see whether the name is in the dataset of names. If so, we compute the probability of the name being male or female based on the number of times that it appears with the given gender: $\textnormal{count}(female)$ or $\textnormal{count}(male)$. If the name is not in the database of names we label it as unknown. The MLE is computed using Equation~\ref{eq:ml_gender}. The probability of the name being male is $p(male)=1-p(female)$. We use a threshold $\tau_g$ to decide on the final label: a name is classified as female if $ML(female) > \tau_g $, male if $ML<1-\tau_g$, and ambiguous otherwise. In our performance evaluation we used three different threshold values: $\tau_g$= 0.50, 0.75, and 0.90.). We developed five different ML models. Four of the models were trained on the following four datasets: DIME, Facebook, Florida, SSA. The fifth model was trained on the combined dataset.

\subsection*{Hybrid Approaches}
\label{sec:hybrid_approaches}
In addition to the MLE approach we also introduce two new hybrid methods combining existing approaches with our MLE approach in order to obtain higher accuracy. 
The first hybrid approach is based on a two-stage process. In the first stage we run our MLE approach. In the second stage, for the names whose MLEs are in the range between 0.25 and 0.75, we use the best model from the existing tools.

The second hybrid approach assigns gender based on the voting across the top three most accurate approaches such that the final gender is assigned based on the majority votes. We then combine this voting approach with our MLE model in a two-stage setting as in the case with the first hybrid approach.

\section*{Performance Evaluation}
\label{sec:gender_performance_evaluation}
We use the six test sets to evaluate the performance of the existing name-to-gender inference approach and our proposed MLE approach. In this section we first describe the evaluation setup and then present the performance evaluation results. 

\subsection*{Evaluation Setup}
\label{sec:gender_evaluation_setup}
We constructed six test sets using a random 10\% subset of the four datasets (DIME, Florida, Facebook, and SSA) and the complete ACL and CMU datasets. We decided to use a subset of the four datasets for three reasons: 1) the size of the four datasets - running all of them through the different approaches would be very time consuming; 2) some of the approaches involve paid services and running all names would incur a high cost; and 3) in order to train our MLE model we had to use a portion of the datasets as the training set. Therefore we used the remaining 90\% of the four datasets as the training set to obtain the parameters of our MLE model. 

Shown in Table~\ref{tab:test_sets} is a summary of the number of unique names across the constructed test sets along with the gender percentage distribution. 

\begin{table}[ht]
    \centering
    \begin{tabular}{|l|r|r|r|r|}\hline
        \bf Dataset        & \bf Sampling Ratio & \bf \# of Unique Names & \bf \% Female & \bf \% Male \\ \hline
        ACL            & 100\%          &  5,244             & 33.73 & 66.27\\ \hline
        CMU            & 100\%          &  7,211             & 64.25 & 35.75\\ \hline
        DIME           & 10\%           &  6,789             & 40.71 & 59.29\\ \hline
        Facebook       & 10\%           &  9,862             & 45.07 & 54.93\\ \hline
        Florida        & 10\%           & 46,904             & 67.78 & 32.22\\ \hline
        SSA           & 10\%           & 53,059             & 64.21 & 35.79\\ \hline
    \end{tabular}
    \caption{Sampling ratio and the number of unique names across the six test sets.}
    \label{tab:test_sets}
\end{table}

\subsection*{Evaluation Results and Analysis}
\label{sec:gender_evaluation_results}
In this section, we provide a summary of the evaluation results and analysis. We start with the performance evaluation results. We proceed by measuring gender bias error across the different approaches. Lastly, we perform error analysis over the names features such as their length and character sequences. 

\subsubsection*{Performance Evaluation Results}
\label{sec:performance comparison results}
Inferring gender from a person's name is essentially a text classification problem. Therefore, we used classification measures to evaluate the various approaches. Performance evaluation analysis was done using the following four evaluation measures: accuracy, precision, recall, and F1 score. We used the female gender as the positive class. In a given test set $k$ for each name $n^k_i$ we have the observing gender $g^k_i$ and the predicted gender $\hat{g}^k_i$. All measures were computed using their standard definitions. For better readability all measures' values are presented as percentages. For example, accuracy is computed by going over all test names $n^k_i$ and counting how many times the predicted gender $\hat{g}^k_i$ is equal to the observed gender $g^k_i$.  We refer to this as $\text{Count}(\hat{g}^k_i==g^k_i)$. Accuracy is the number of correctly labeled names divided by the total number of names in the test set. In other words, the accuracy is the proportion of correctly labeled names out of all test names $\text{Count}(g^k_i)$: 
\begin{eqnarray} 
\label{eq:accuracyk} 
Accuracy^k=\frac{Count(\hat{g}^k_i==g^k_i)}{Count(g^k_i)}
\end{eqnarray}

We multiply this value by 100 to convert it into a percentage value. While the latter three measures (precision, recall, and F1) are typically used in instances when the dataset is highly imbalanced, across our test sets this is not the case. However, we decided to include these measures in order to provide a more detailed analysis which would be helpful when the gender inference approaches are used on precision or recall oriented tasks. In other words, having these measures help us compare the approaches on tasks when having many false female names (false positives) may impose a high cost (i.e. a precision oriented task) or when having many false male names (false negatives) may incur a high cost (i.e. recall oriented task). 

Approaches were evaluated across the six test sets and the final performance evaluation was performed on the aggregated scores. Scores were aggregated as a weighted average. For example, the overall accuracy of an approach across the $k$ test sets, where each test set contains $s_k$ names, was computed as:
\begin{eqnarray} 
\label{eq:accuracy} 
Accuracy=\frac{\sum_k{s_kAccuracy^k}}{\sum_k{s_k}}
\end{eqnarray}

Shown in Table~\ref{tab:gender_performance_evaluation_average} are the overall acuracy, precision, recall, and F1 score across all gender inference approaches. 
For each of the four evaluation measures we provide results over the individual datasets in Appendices A (accuracy), B (Precision), C (Recall), and D (F1). 

\begin{table*}[ht]
\centering
\begin{tabular}{|l|r|r|r|r|r|r|r|}\hline
\bf Inference Approach		& \bf Accuracy 	& \bf Precision	& \bf Recall & \bf F1	 & \bf GBE   \\\hline
Baby Name Guesser		& 61.07	    & 70.9 	    & 63.44	 & 66.31 &  -7.82 \\\hline
chicksexer (CS)		    & 92.28	    & 92.67	    & 93.54	 & 93.01 &   0.26 \\\hline
Gender API		        & 76.97	    & 84.65	    & 75.69	 & 79.47 &  -6.72 \\\hline
gender-guesser		    & 26.77	    & 34.52	    & 25.89	 & 28.44 & -20.99 \\\hline
gender R		    & 58.12	    & 67.10	    & 60.58	 & 62.56 &  -6.97 \\\hline
NamSor		            & 90.48	    & 93.31	    & 89.93	 & 91.54 &	-2.48 \\\hline
MLE DIME		        & 58.18	    & 70.66	    & 56.75	 & 62.30 & -13.70 \\\hline
MLE Facebook		    & 26.37	    & 37.20	    & 26.44	 & 30.06 & -20.47 \\\hline
MLE Florida		        & 73.87	    & 74.28	    & 82.17	 & 77.57 &   6.08 \\\hline
MLE SSA		            & 52.45	    & 61.84	    & 56.38	 & 57.54 &  -6.72 \\\hline
MLE All (MLEA)		& 72.67	    & 77.18	    & 79.80	 & 77.89 &	 1.23 \\\hline
MLEA+CS		            & 93.35	    & 93.81	    & 95.11	 & 94.37 &	 0.58 \\\hline
MLEA+voting 		    & 92.67	    & 94.42	    & 93.83	 & 93.80 &  -0.67 \\\hline
\end{tabular}
\caption{Average accuracy, precision, recall, and F1 computed over various name-to-gender inference approaches. Approaches were evaluated across six different datasets. Shown in the rightmost column is the gender bias error (GBE).}
\label{tab:gender_performance_evaluation_average}
\end{table*}

All MLE models were run with a threshold value of $\tau_{g}$= 0.90. Across all the measures there are three consistent top performers from the existing gender inference approaches: chicksexer (CS), Namsor, and Gender API. The CS approach is consistently the top one except for precision, in which Namsor is the leader. 

Across the MLE models, the most accurate ones are the MLE Florida and the model that was trained on the combined data - "MLE All" (MLEA), with the latter one having a higher F1 score. 

We also evaluated our first hybrid approach by combining the MLEA model and the best model from the existing tools which is the CS model. In the first stage we ran the MLEA model and we used the CS model in the second stage of this approach. We refer to this approach as the "MLEA+CS".

In addition, we ran the second hybrid approach using the top three most accurate approaches (i.e. CS, Namsor, and Gender API) and the MLEA approach. This approach assigns gender based on the voting across the top three approaches. It then combines this voting approach with our MLEA model in a two-stage setting as in the case with the "MLEA+CS" approach. We refer to this model as the "MLEA+voting". 

Based on the accuracy and the F1 score we concluded that across all approaches the "MLA+CS" model yields the best results throughout all datasets with an average accuracy of $93.35$ and an average F1 score of $94.37$. 

\subsubsection*{Gender Bias Error}
\label{sec:gender_bias_error}
Algorithmic bias is defined as systematic and repeatable errors that create unfair outcomes. While precision and recall emphasize the importance of having fewer false female and false male names, they do not directly answer the question of whether the algorithmic approach is biased towards one particular outcome (i.e. inferred gender). We used the gender bias error (GBE) \cite{wais_2016} as a measure for the algorithmic bias present in a particular gender inference approach. This measure is defined as the difference between the number of false female (FF) and false male (FM) names normalized by the total number of names which is the sum of the false and true inferred names:
\begin{eqnarray} 
\label{eq:gbe} 
GBE=\frac{FF-FM}{FF+FM+TF+TM}
\end{eqnarray}
where TF is the number of true female names and TM is the number of true male names. This measure uses the sign as an indicator of the bias direction. A negative sign indicates that more female names are classified as male, thereby underestimating the proportion of female names in the collection. On the other hand, a positive sign indicates that more male names are classified as female, thus overestimating the proportion of female names. Ideally, the inference approach should have a GBE close to or equal to zero, which indicates an even number of false female and false male names. Shown in the rightmost column of Table~\ref{tab:gender_performance_evaluation_average} are the GBE values across the different approaches. As is the case with the four performance measures, this is the aggregated GBE score computed over the six test sets as a weighted average. We give the GBE values across the individual test sets in Appendix~\ref{appendix:individual_gbe}. The top three approaches with GBE values closest to zero are: CS, MLEA+CS, and MLEA+voting. 

\subsubsection*{Error Analysis}
\label{sec:error_analysis}
We performed error analysis by examining the characteristics of the inference approach input, which is a person's name, across the four rate types: TF, TM, FF, and FM. Given that a person's name is a sequence of characters, our analysis focused on the length of the name (i.e. the number of characters), character type, and the attributes of the character sequence. Character type analysis were performed over names that contain at least one character that is not part of the English alphabet. We refer to these as non-English names (e.g, Špela, Özgür, etc.). We examined the performance of the inference approaches over these name types.  

We examined the character sequence attributes with a widely used family of language models called N-grams. While N-grams typically operate on a word level and are used to compute the probability of having a particular sentence or sequence of words, in our study we applied this model on a character level. This family of language models predicts the probability of the next word in the sentence using the previous N-1 words. We used the bigram and trigram versions, which compute the probability of the next character given the previous character (bigram) and the previous two characters (trigram). Shown in the subsequent subsections are the error analysis results.  

\subsubsection*{Name Length}
\label{sec:name_length}
\begin{figure}[!ht]
   \centering
   \includegraphics[width=0.60\linewidth]{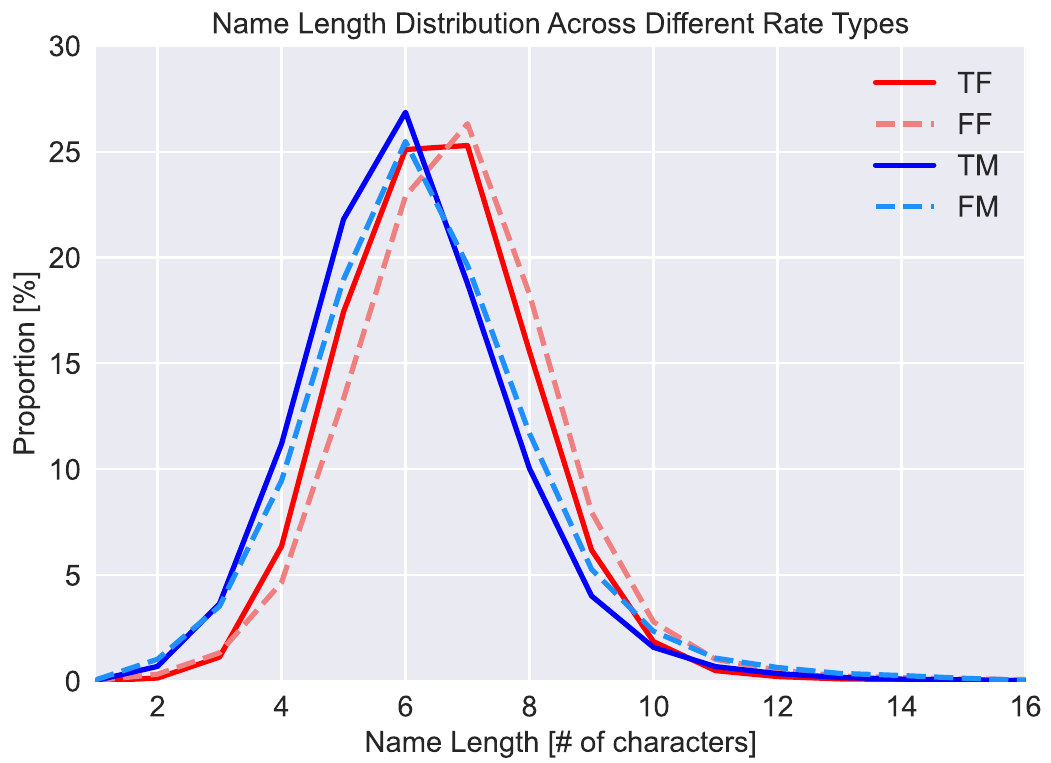}
   \caption{Percentage of names distribution across character length computed over four rate types: true female (TF), false female (FF), true male (TM), and false male (FM).}
   \label{fig:nl_distribution}
\end{figure}

A conveniently measurable characteristic of a person's name is its length (i.e., the number of characters present in the name). We used this characteristic to perform error analysis based on the input name character length. Across all inference approaches we went over the gender inference results and measured the character length of the names belonging to each of the four rate types. For each rate type, we computed a histogram over the character lengths of the names belonging to that rate type. A summary of the analysis is provided in Fig~\ref{fig:nl_distribution}, where the x-axis is the name character length and the y-axis is the percentage of names within the rate type with the specific character length. From the four histograms, one could immediately notice that the name length of FM and FF names are skewed towards higher character lengths compared to the TM and TF names. This implies that on average across all gender inference types, names with longer character lengths have a higher likelihood of being mislabeled.

\subsubsection*{Character Type Analysis}
\label{sec:character_type}
\begin{figure}[!ht]
   \centering
   \includegraphics[width=0.60\linewidth]{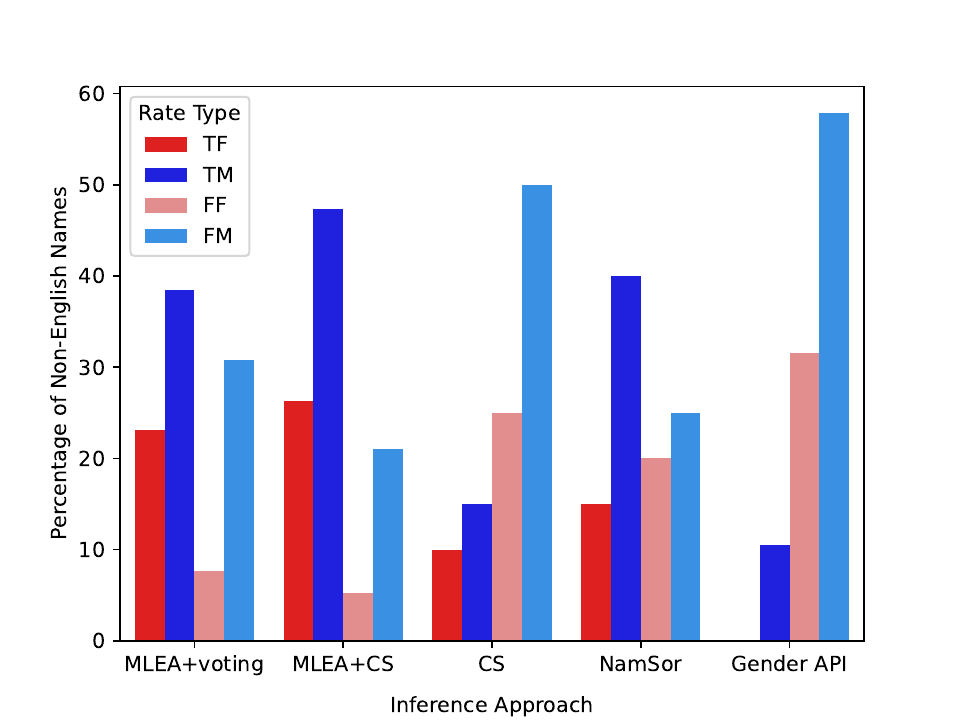}
   \caption{Distribution of non-English names (in percentages) across the four rate types. Analysis were performed over the top five most accurate inference approaches.}
   \label{fig:non_english_analysis}
\end{figure}

Person names can contain at least one character which is not part of the English alphabet. As stated earlier, we refer to these as non-English names. In order to determine how sensitive the inference approaches are to non-English names, we computed their percentage distribution across the four rate types. In Fig~\ref{fig:non_english_analysis} we show the percentage distribution for the top five most accurate inference approaches.  

For MLEA+voting, MLEA+CS, and NamSor, we observe that the percentage of non-English TF and TM names is higher than FF and FM, except for NamSor where non-English FF names are higher than TF. We also observe that for CS and Gender API the percentage of non-English FF and FM is higher than TF and TM. This suggests that both of these inference approaches are more sensitive to non-English names than the other approaches. In fact, these analyses demonstrate that combining MLEA and CS helps improve the performance over non-English names, as the percentage ratio between true and false rate types is reversed. Lastly, we observe that the percentage of non-English TF names for Gender API is zero.

\subsubsection*{Character Sequence Attributes}
\label{sec:character_sequence_attributes}
\begin{figure}[!ht]
   \centering
   \includegraphics[width=0.60\linewidth]{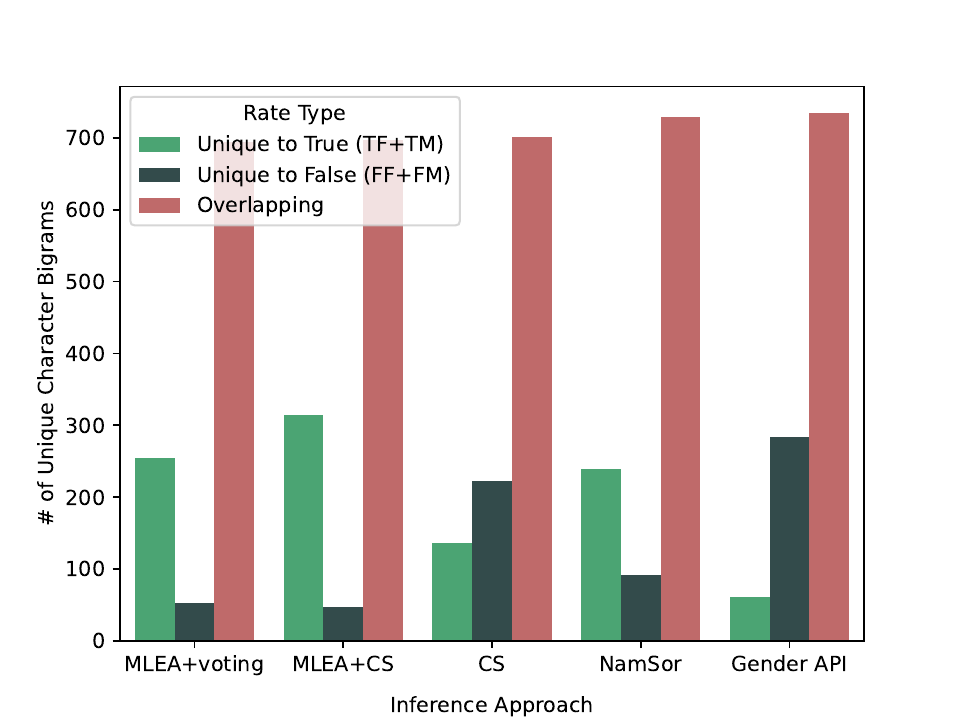}
   \caption{Number of unique and overlapping bigrams across true and false inferred genders. Analysis were performed over the top five most accurate inference approaches.}
   \label{fig:unique_bigrams}
\end{figure}

\begin{figure}[!ht]
   \centering
   \includegraphics[width=0.60\linewidth]{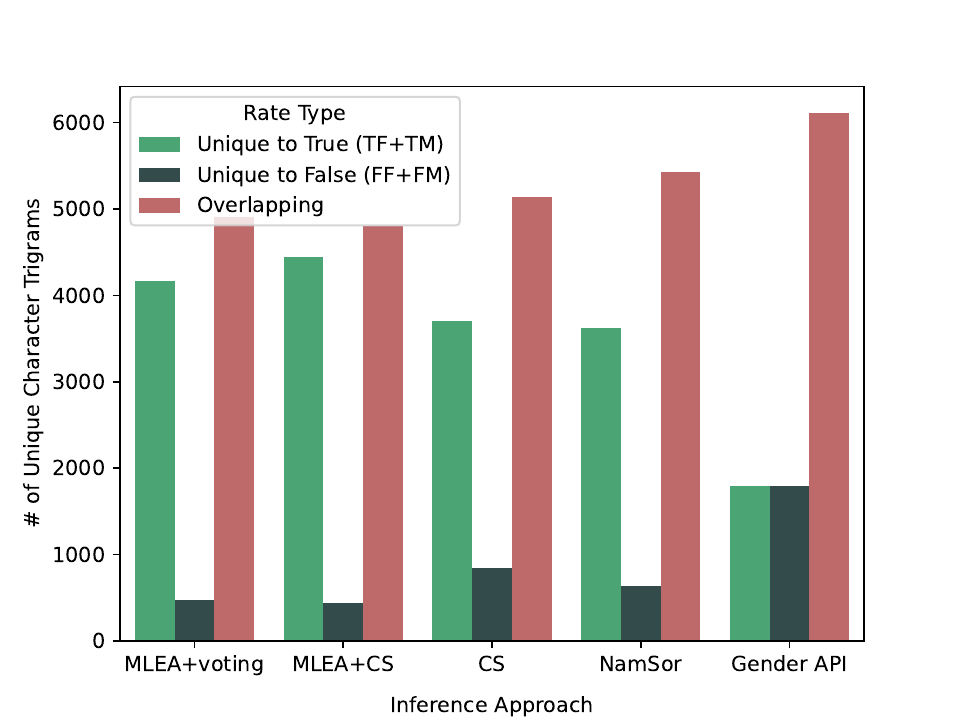}
   \caption{Number of unique and overlapping trigrams across true and false inferred genders. Analysis were performed over the top five most accurate inference approaches.}
   \label{fig:unique_trigrams}
\end{figure}

Across all inference approaches and rate types we extracted all possible character bigrams and trigrams that are contained within a person's name. For example, if a person's name is "Mary" we would extract the following three bigrams: "Ma", "ar", and "ry" and the following two trigrams: "mar" and "ary". These character bigrams and trigrams were used as person's name attributes with which we analyzed the four rate types. Rate types were combined into two categories: 1) true - containing TF and TM, and 2) false - containing FF and FM. Across the two categories we determined the number of unique and overlapping bigrams and trigrams. The N-gram analysis were performed over the five best performing systems as measured by the F-1 score. Figs~\ref{fig:unique_bigrams} and \ref{fig:unique_trigrams} display results of our bigram and trigram analysis. 

The results of these analyses suggest that further gender inference improvements could be achieved by considering the unique bigrams and trigrams contained within the false rate types (FF and FM). We note how the number of unique bigrams contained within false inferred names is notably higher than in true names for the CS and Gender API approaches. This means that both of these approaches fail to adequately take advantage of the unique bigrams contained within a person's name. 
Ideally, we would like to have the number of unique bigrams and trigrams contained within FF and FM names be close to or equal to zero. In case of the trigram results, we note that except for Gender API where the number of trigrams unique to true and false inferred names is almost identical, the remaining inference approaches take advantage of the unique trigrams. Nevertheless, further accuracy improvements could be made if we focus on the unique true trigrams.

\section*{Conclusion}
\label{sec:conclusion}
We performed a large scale performance evaluation study of existing name-to-gender inference approaches. We evaluated the performance of six existing approaches.  Furthermore, we proposed our own gender inference approach, which is based on maximum likelihood estimates and is straightforward to implement. In addition, we proposed two methods for combining the results of individual inference approaches. Using multiple different evaluation measures, we concluded that a combination of our MLEA approach and the top performing existing approach (MLEA+CS) provides the most accurate prediction. Inference approaches were also evaluated on the level of bias of their algorithms using the gender bias error (GBE) measure. We concluded that the CS approach is least biased, followed by the MLEA+CS. Finally, through a detailed analysis of the input names and their character sequence attributes, we characterized the errors across all inference approaches and individually for the top five best performing systems.  
\section*{Acknowledgements}
\label{sec:acknowledgements}
The authors would like to thank Gender API and NamSor for providing us with free access to their APIs for the duration of this study. We would also like to thank Rui Bai, Jillian Ross, Zhouyao Xie, and Jialu Xia for their significant contributions in the creation of the datasets and evaluation of the inference approaches. Lastly, the authors would like to thank the Data Science Institue at Columbia University for the generous funding support. 

\bibliography{plos_gender}

\clearpage

\appendix

\section{Accuracy Results Across Individual Test Sets}
\label{appendix:individual_accuracy}
Shown in Table~\ref{tab:accuracy_individual} are accuracy results computed over the six individual test sets.

\begin{table*}[ht]
\centering
\begin{tabular}{|l|r|r|r|r|r|r|r|}\hline
\bf Inference Approach      & \bf ACL   & \bf CMU   & \bf DIME  & \bf Facebook & \bf Florida & \bf SSA   & \bf Agg. \\\hline
Baby Name Guesser	    & 82.82	& 78.82	& 64.06	& 79.43	   & 49.32	 & 63.08 & 61.06     \\\hline
chicksexer (CS)		    & 80.15	& 96.2	& 81.27	& 86.96	   & 88.83	 & 98.37 & 92.27     \\\hline
Gender API		        & 49.34	& 76.7	& 74.74	& 74.41	   & 70.42	 & 86.28 & 76.96     \\\hline
gender-guesser		    & 54.54	& 56.12	& 43.22	& 53.35	   & 19.26	 & 19.62 & 26.76     \\\hline
gender R		    & 50.01	& 80.65	& 60.51	& 75.74	   & 51.51	 & NA	 & 58.12     \\\hline
NamSor		            & 84.31	& 93.94	& 81.62	& 88.82	   & 88.01	 & 94.23 & 90.47	 \\\hline
MLE DIME		        & 57.52	& 81.49	& NA	& 76.37	   & 44.6	 & 63.68 & 58.17     \\\hline
MLE Facebook		    & 42.57	& 52.85	& 40.08	& NA	   & 21.51	 & 23.70 & 26.36     \\\hline
MLE Florida		        & 57.16	& 79.99	& 55.87	& 72.95	   & NA	     & 77.15 & 73.86	 \\\hline
MLE SSA		            & 44.12	& 75.4	& 54.81	& 67.98	   & 46.23	 & NA	 & 52.44   	 \\\hline
MLE All (MLEA)          & NA	& NA	& 55.32	& 74.19	   & 58.71	 & 86.94 & 72.67  	 \\\hline
MLEA+CS		            & NA	& NA	& 81.37	& 89.55	   & 89.76	 & 98.76 & 93.35  	 \\\hline
MLEA+voting             & NA	& NA	& 81.20	& 90.01	   & 88.72	 & 98.12 & 92.67  	 \\\hline
\end{tabular}
\caption{Accuracy over individual datasets and aggregate accuracy (rightmost columns) computed as a weighted average.}
\label{tab:accuracy_individual}
\end{table*}

\section{Precision Results Across Individual Test Sets}
\label{appendix:individual_precision}
Shown in Table~\ref{tab:precision_individual} are precision results computed over the six individual test sets.

\begin{table*}[ht]
\centering
\begin{tabular}{|l|r|r|r|r|r|r|r|}\hline
\bf Inference Approach      & \bf ACL   & \bf CMU   & \bf DIME  & \bf Facebook & \bf Florida & \bf SSA   & \bf Agg. \\\hline
Baby Name Guesser    & 71.44 & 85.36 & 54.19 & 71.48    & 67.52   & 73.88 & 70.9      \\ \hline
chicksexer (CS)      & 68.27 & 97.7  & 73.58 & 79.74    & 93.41   & 98.57 & 92.67     \\ \hline
Gender API           & 33.84 & 87.27 & 67.01 & 67.99    & 86.44   & 93.08 & 84.65     \\ \hline
gender-guesser       & 38.62 & 70.74 & 36.81 & 48.56    & 31.62   & 28.85 & 54.67     \\ \hline
gender R         & 35.42 & 88.00 & 51.01 & 68.02    & 69.56   & NA    & 67.10     \\ \hline
NamSor               & 76.68 & 96.96 & 74.77 & 84.93    & 93.94   & 97.83 & 93.31     \\ \hline
MLE DIME             & 40.83 & 91.29 & NA    & 70.18    & 64.26   & 76.55 & 70.66     \\ \hline
MLE Facebook         & 29.06 & 67.05 & 34.76 & NA       & 35.95   & 35.35 & 37.2      \\ \hline
MLE Florida          & 41.31 & 85.99 & 47.25 & 64.21    & NA      & 81.27 & 74.28     \\ \hline
MLE SSA              & 30.60 & 84.19 & 46.40 & 60.63    & 64.37   & NA    & 61.84     \\ \hline
MLE All (MLEA)       & NA    & NA    & 46.91 & 64.81    & 72.77   & 87.24 & 77.18     \\ \hline
MLEA+CS              & NA    & NA    & 73.18 & 82.22    & 98.86   & 98.86 & 93.81     \\ \hline
MLEA+voting          & NA    & NA    & 75.08 & 83.98    & 93.65   & 98.84 & 94.42     \\ \hline
\end{tabular}
\caption{Precision over individual datasets and aggregate precision (rightmost column) computed as a weighted average.}
\label{tab:precision_individual}
\end{table*}
\clearpage

\section{Recall Results Across Individual Test Sets}
\label{appendix:individual_recall}
Shown in Table~\ref{tab:recall_individual} are recall results computed over the six individual test sets.

\begin{table*}[ht]
\centering
\begin{tabular}{|l|r|r|r|r|r|r|r|}\hline
\bf Inference Approach      & \bf ACL   & \bf CMU   & \bf DIME  & \bf Facebook & \bf Florida & \bf SSA   & \bf Agg. \\\hline
Baby Name Guesser    & 71.44 & 85.36 & 54.19 & 71.48    & 67.52   & 73.88 & 70.9      \\ \hline
chicksexer (CS)      & 68.27 & 97.7  & 73.58 & 79.74    & 93.41   & 98.57 & 92.67     \\ \hline
Gender API           & 33.84 & 87.27 & 67.01 & 67.99    & 86.44   & 93.08 & 84.65     \\ \hline
gender-guesser       & 38.62 & 70.74 & 36.81 & 48.56    & 31.62   & 28.85 & 54.67     \\ \hline
gender R         & 35.42 & 88.00 & 51.01 & 68.02    & 69.56   & NA    & 67.10     \\ \hline
NamSor               & 76.68 & 96.96 & 74.77 & 84.93    & 93.94   & 97.83 & 93.31     \\ \hline
MLE DIME             & 40.83 & 91.29 & NA    & 70.18    & 64.26   & 76.55 & 70.66     \\ \hline
MLE Facebook         & 29.06 & 67.05 & 34.76 & NA       & 35.95   & 35.35 & 37.2      \\ \hline
MLE Florida          & 41.31 & 85.99 & 47.25 & 64.21    & NA      & 81.27 & 74.28     \\ \hline
MLE SSA              & 30.60 & 84.19 & 46.40 & 60.63    & 64.37   & NA    & 61.84     \\ \hline
MLE All (MLEA)       & NA    & NA    & 46.91 & 64.81    & 72.77   & 87.24 & 77.18     \\ \hline
MLEA+CS              & NA    & NA    & 73.18 & 82.22    & 98.86   & 98.86 & 93.81     \\ \hline
MLEA+voting          & NA    & NA    & 75.08 & 83.98    & 93.65   & 98.84 & 94.42     \\ \hline
\end{tabular}
\caption{Recall over individual datasets and aggregate recall (rightmost column) computed as a weighted average.}
\label{tab:recall_individual}
\end{table*}

\section{F1 Results Across Individual Test Sets}
\label{appendix:individual_f1}
Shown in Table~\ref{tab:f1_individual} are F1 results computed over the six individual test sets.

\begin{table*}[ht]
\centering
\begin{tabular}{|l|r|r|r|r|r|r|r|}\hline
\bf Inference Approach      & \bf ACL   & \bf CMU   & \bf DIME  & \bf Facebook & \bf Florida & \bf SSA   & \bf Agg. \\\hline
Baby Name Guesser	  & 76.27 & 83.08	& 63.17	& 79.86    & 56.52	 & 69.58  & 66.31      \\\hline
chicksexer (CS)		  & 72.35 & 97.02	& 78.56	& 86.82    & 91.60	 & 98.73  & 93.01      \\\hline
Gender API		      & 41.18 & 80.44	& 70.67	& 74.21    & 75.39	 & 88.82  & 79.47      \\\hline
gender-guesser		  & 46.69 & 61.29	& 44.13	& 53.38    & 21.63	 & 21.54  & 28.44      \\\hline
gender R		  & 44.12 &	84.31	& 60.89	& 76.41    & 58.60	 & NA	  & 62.56      \\\hline
NamSor		          & 76.78 &	95.20	& 78.57	& 88.06    & 90.86	 & 95.40  & 91.54      \\\hline
MLE DIME		      & 47.78 &	84.53	& NA	& 75.94    & 50.17	 & 68.89  & 62.30      \\\hline
MLE Facebook		  & 36.40 &	58.78	& 42.23	& NA       & 25.87	 & 27.66  & 30.06      \\\hline
MLE Florida		      & 50.23 &	84.08	& 57.11	& 75.06	   & NA	     & 82.47  & 77.57      \\\hline
MLE SSA		          & 38.46 &	79.87	& 56.06	& 69.92	   & 53.85	 & NA     & 57.54      \\\hline
MLE All (MLEA)		  & NA	  & NA	    & 57.42	& 76.56	   & 67.22	 & 90.17  & 77.89      \\\hline
MLEA+CS	  	          & NA	  & NA	    & 78.90	& 89.43	   & 92.36	 & 99.03  & 94.37      \\\hline
MLEA+voting 	 	  & NA	  & NA	    & 78.37	& 89.80    & 91.49	 & 98.54  & 93.80      \\\hline
\end{tabular}
\caption{F1 over individual datasets and aggregate F1 (rightmost column) computed as a weighted average.}
\label{tab:f1_individual}
\end{table*}

\clearpage
\section{GBE Results Across Individual Test Sets}
\label{appendix:individual_gbe}
Shown in Table~\ref{tab:bias_individual} are GBE results computed over the six individual test sets.

\begin{table*}[ht]
\centering
\begin{tabular}{|l|r|r|r|r|r|r|r|}\hline
\bf Inference Approach      & \bf ACL   & \bf CMU   & \bf DIME  & \bf Facebook & \bf Florida & \bf SSA   & \bf Agg. \\\hline
Baby Name Guesser		&  4.90 &  -3.34 & 20.19  & 11.97	 & -18.99  &  -7.06 &  -7.81	\\\hline
chicksexer (CS)		    &  4.29	&  -0.89 &  5.91  &  8.78	 &  -2.58  &   0.21	&   0.25	\\\hline
Gender API		        & 18.69	&  -9.32 &  4.70  &  9.08	 & -15.37  &  -5.61	&  -6.71	\\\hline
gender-guesser		    & 17.81	& -15.14 & 16.17  &  9.94	 & -32.45  & -25.97	& -20.98	\\\hline
gender R		    & 21.96	&  -5.17 & 19.55  & 12.68	 & -18.45  &	NA	&  -6.97    \\\hline
NamSor		            &  0.10	&  -2.29 &  4.36  &  3.45	 &  -4.30  &  -3.12	&  -2.47	\\\hline
MLE DIME		        & 13.84	&  -8.86 &	NA	  &  8.05	 & -24.38  & -11.67	& -13.69    \\\hline
MLE Facebook		    & 22.80	& -14.12 & 22.30  &	  NA	 & -29.69  & -22.93	& -20.47    \\\hline
MLE Florida		        & 18.59	&  -2.79 & 21.48  &	17.68	 &    NA   &   1.91	&   6.07	\\\hline
MLE SSA		            & 23.32	&  -6.27 & 21.42  &	16.30	 & -19.05  &   NA	&  -6.71    \\\hline
MLE All (MLEA)		    & NA	& NA	 & 23.49  &	19.97	 &  -9.61  &   4.47	&   1.22	\\\hline
MLEA+CS		            & NA	& NA	 &  6.91  &	 8.66	 &  -1.65  &   0.23	&   0.57	\\\hline
MLEA+voting 		    & NA	& NA	 &  4.65  &  6.82	 &  -2.83  &  -0.82	&  -0.66	\\\hline
\end{tabular}
\caption{GBE over individual datasets and aggregate GBE (rightmost column) computed as a weighted average.}
\label{tab:bias_individual}
\end{table*}

\end{document}